\documentclass[letterpaper, 10 pt, conference]{ieeeconf}  

\IEEEoverridecommandlockouts                       

\usepackage{amsmath}
\usepackage{xcolor, soul}
\usepackage{subcaption}
\usepackage{amsfonts}
\usepackage{amssymb}
\usepackage{bm}
\usepackage{multicol}
\usepackage{multirow}
\usepackage{graphics, graphicx}
\usepackage{scalerel}
\usepackage{tikz}
\usetikzlibrary{svg.path}

\usepackage[compact]{titlesec}
\titlespacing{\section}{1pt}{*1}{*1}
\titlespacing{\subsection}{1pt}{*1}{*1}

\definecolor{orcidlogocol}{HTML}{A6CE39}
\tikzset{
  orcidlogo/.pic={
    \fill[orcidlogocol] svg{M256,128c0,70.7-57.3,128-128,128C57.3,256,0,198.7,0,128C0,57.3,57.3,0,128,0C198.7,0,256,57.3,256,128z};
    \fill[white] svg{M86.3,186.2H70.9V79.1h15.4v48.4V186.2z}
                 svg{M108.9,79.1h41.6c39.6,0,57,28.3,57,53.6c0,27.5-21.5,53.6-56.8,53.6h-41.8V79.1z M124.3,172.4h24.5c34.9,0,42.9-26.5,42.9-39.7c0-21.5-13.7-39.7-43.7-39.7h-23.7V172.4z}
                 svg{M88.7,56.8c0,5.5-4.5,10.1-10.1,10.1c-5.6,0-10.1-4.6-10.1-10.1c0-5.6,4.5-10.1,10.1-10.1C84.2,46.7,88.7,51.3,88.7,56.8z};
  }
}

\newcommand\orcidiconKLW[1]{\href{https://orcid.org/0000-0002-1938-4222}{\mbox{\scalerel*{
\begin{tikzpicture}[yscale=-1,transform shape]
\pic{orcidlogo};
\end{tikzpicture}
}{|}}}}
\newcommand\orcidiconAK[1]{\href{https://orcid.org/0000-0003-1067-1134}{\mbox{\scalerel*{
\begin{tikzpicture}[yscale=-1,transform shape]
\pic{orcidlogo};
\end{tikzpicture}
}{|}}}}
\newcommand\orcidiconFGS[1]{\href{https://orcid.org/0000-0002-5090-9007}{\mbox{\scalerel*{
\begin{tikzpicture}[yscale=-1,transform shape]
\pic{orcidlogo};
\end{tikzpicture}
}{|}}}}

\usepackage{hyperref}

\title{\LARGE \bf
Model Predictive Wave Disturbance Rejection for Underwater Soft Robotic Manipulators
}

\author{Kyle L. Walker$^{1,3}$\orcidiconKLW{0000-0002-1938-4222}, Cosimo Della Santina$^{2}$\orcidiconFGS{0000-0003-1067-1134} and Francesco Giorgio-Serchi$^{1}$\orcidiconFGS{0000-0002-5090-9007}
\thanks{This work was supported by the EPSRC under grant No. EP/R026173/1 and grant No. EP/R513209/1. This work was also supported by the Scottish Funding Council (SFC) through the SFC Saltire Emerging Researcher Scheme (Scottish Universities Physics Alliance (SUPA) European Exchanges) under SFC Project Code: H22054.}
\thanks{$^{1}$Kyle L. Walker and Francesco Giorgio-Serchi are with the Institute for Integrated Micro and Nano Systems, University of Edinburgh, Edinburgh, U.K.. Correspondence: f.giorgio-serchi@ed.ac.uk}%
\thanks{$^{2}$Cosimo Della Santina is with the Cognitive Robotics Department, TU Delft, Delft, The Netherlands.}
\thanks{$^{1}$Kyle L. Walker is also with the National Robotarium, Boundary Road North, Heriot Watt University, Edinburgh, U.K.}}%

\begin{document}

\maketitle
\thispagestyle{empty}
\pagestyle{empty}

\begin{abstract}

Inspired by the octopus and other animals living in water, soft robots should naturally lend themselves to underwater operations, as supported by encouraging validations in deep water scenarios. This work deals with equipping soft arms with the intelligence necessary to move precisely in wave-dominated environments, such as shallow waters where marine renewable devices are located. This scenario is substantially more challenging than calm deep water since, at low operational depths, hydrodynamic wave disturbances can represent a significant impediment. We propose a control strategy based on Nonlinear Model Predictive Control that can account for wave disturbances explicitly, optimising control actions by considering an estimate of oncoming hydrodynamic loads. The proposed strategy is validated through a set of tasks covering set-point regulation, trajectory tracking and mechanical failure compensation, all under a broad range of varying significant wave heights and peak spectral periods. The proposed control methodology displays positional error reductions as large as 84\% with respect to a baseline controller, proving the effectiveness of the method. These initial findings present a first step in the development and deployment of soft manipulators for performing tasks in hazardous water environments.

\end{abstract}

\section{INTRODUCTION}


Soft robotics has seen a surge in recent times \cite{Aracri2021}, as their soft structure, adaptability and compliance are all highly attractive features, particularly within the context of a marine environment where fluid loading can be considerable \cite{Trivedi2008, Renda2014}. Several works have exploited these characteristics for varying purposes underwater, such as for marine sampling where handling delicate organisms requires compliant interactions during manipulation \cite{Galloway2016, Gong2021}, or for achieving locomotion through inspiration from aquatic organisms \cite{Calisti2017,GiorgioSerchi2017}. In terms of manipulation, consideration of external loading is required to achieve accurate and reliable performance; this is even more relevant when the manipulator is deployed on a floating base, as an unwanted response can be substantial. Here we consider a fixed base, but the translation to deployment on a floating base is achievable through modifications to consider a varying point of evaluation.

Modelling of external loads has been proposed using various approaches \cite{Renda2014, Armanini2021}, but rarely do these works consider the presence of an active fluid. Particularly for applications in an ocean environment, effects from surface waves and subsurface currents can be significant, exposing the robot to potential large-magnitude displacements and therefore requiring mitigation techniques to be considered. Amongst those proposed are the use of state observers \cite{Dian2022} or utilising model-based control \cite{DellaSantina2021}, for example Model Predictive Control (MPC) \cite{Best2016, Bruder2019, Hyatt2020}. The latter is a promising approach, as it allows the inclusion of a disturbance model to be embedded within the control scheme; the controller can therefore explicitly account for perturbations during the optimisation phase. This concept has previously been proposed for the control of subsea vehicles \cite{Fernandez2017, Walker2021_RAL, WalkerIROS}, so a similar approach for a soft manipulator could aid in significantly improving end-effector control. Applying the strategy to real-time, real-world operations becomes reliant on the ability to estimate the robot state and perform the optimisation quickly, with several solutions proposed for both of these within the literature \cite{Loo2019, Hughes2021, Hu2023}. 

With regards to this, we propose deploying MPC for the purpose of controlling a multi-segment continuum soft robot under the influence of unsteady ocean wave disturbances. The fully-coupled fluid-solid dynamical representation of the robot is presented according to a Piece-wise Constant Curvature (PCC) model \cite{Webster2010}, detailing our approach to approximating fluid disturbances across each segment. Two controllers are compared: a Model-Based Kinematic Controller as a baseline and an MPC. The control is tested under three typical scenarios commonly encountered in underwater manipulation tasks: (a) set-point regulation under different degrees of curvature, depths and magnitude of hydrodynamic disturbance; (b) the ability of the controller to follow a prescribed trajectory while subjected to disturbances and (c) failure mitigation, i.e. the capability of the controller to compensate for localized failure of the actuation. In all tests a significant improvement was apparent in contrast to the Kinematic Controller, presenting evidence that a predictive disturbance rejection strategy is key in these contexts to effectively improve control under wave loading. 

This work contributes to the state of the art in marine applications of soft robotics by proposing a control architecture that directly accounts for explicitly modelled time-varying hydrodynamic disturbances. An in-depth analysis of performance under a variety of different real-world wave disturbances, showing that this architecture can autonomously attain a posture and reduce the effect of disturbances on its body throughout.

\section{Manipulator Modelling}

This section presents a modeling framework for a multi-segment soft robot operating underwater in proximity with the free surface. The model accounts for free surface elevation as a result of coincident wave trains which approach the manipulator along the surge (Section \ref{hydro_modelling}), enabling the dynamics to be simplified to that of a planar case. The formulation presented is agnostic to the specifics of the actuator, aiming to maintain generality over the control approach.

\subsection{Kinematics and Dynamics}

\begin{figure}[t!]
    \centering
    \includegraphics[width=0.9\linewidth]{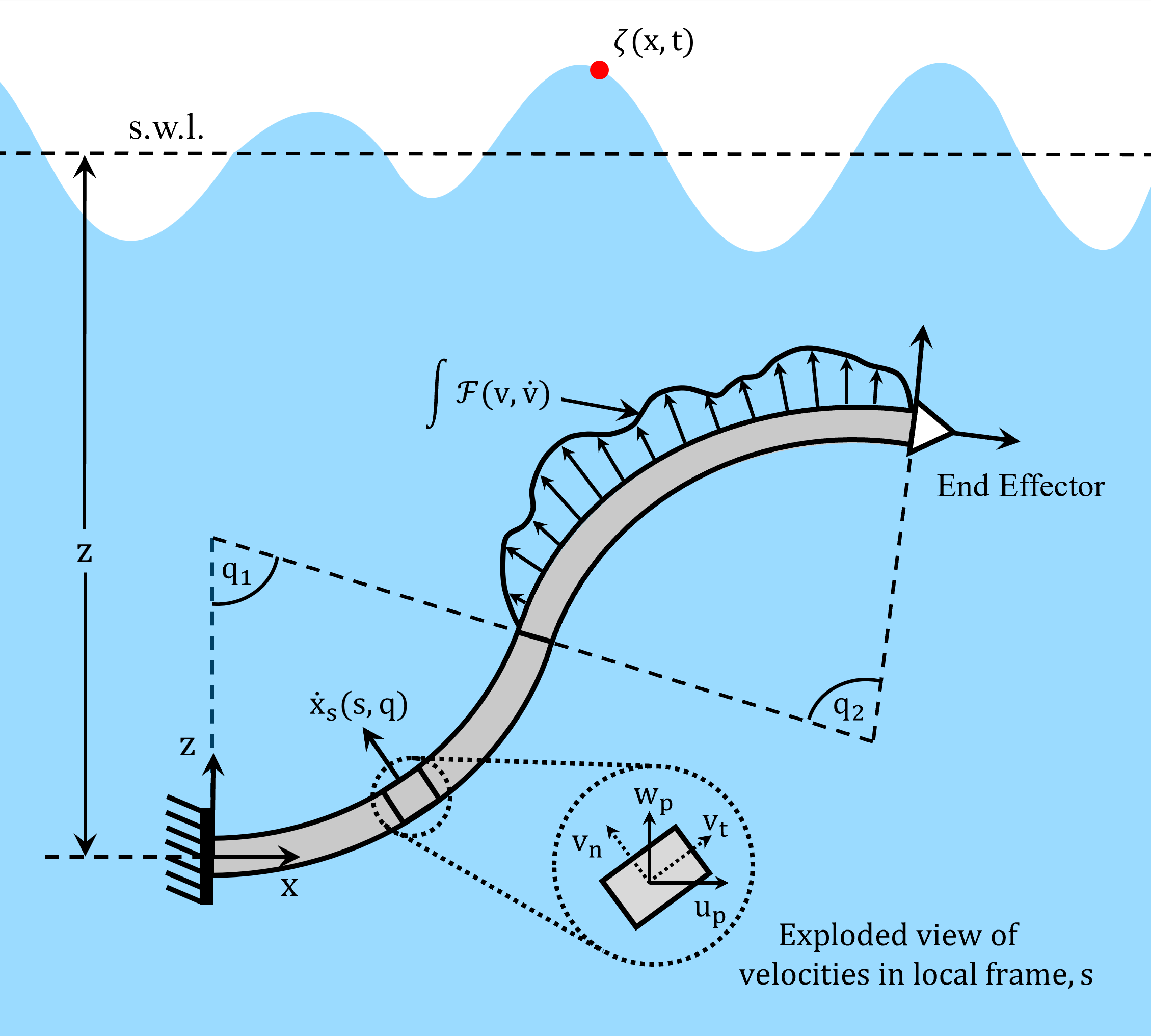}
    \caption{Piece-wise integration of fluid forcing across the segment body, where the relative motion of the fluid is considered to account for non-steady flows and evaluation is performed at each local point, $s$, along the manipulator. }
    \label{fluid_forcing}
\end{figure}

Considering a soft robot with $n$ individually actuated segments such that it is kinematically represented in the configuration space as a vector of joint angles (see Fig. \ref{fluid_forcing}), $q \in \mathbb{R}^{n}$, the position of any point along the manipulator length can be evaluated in the task-space ($\mathbf{x} \in \mathbb{R}^{2}$) according to
\begin{equation} \label{forward_kinematics}
    \mathbf{x}(s,t) = h(s,q(t))
\end{equation}
where $s \in [0,1]$ is the point of evaluation along each segment from base to tip and $h(s,q(t))$ is the forward kinematics at time $t$. Similarly, the Cartesian velocity of each point is described by:
\begin{equation}
    \mathbf{\dot{x}}(s,q) = J(s,q)\dot{q}
\end{equation}
where $J(s,q)=\partial{h(s,q)}/\partial{q}$ is the Jacobian. Given this kinematic representation, the dynamic equation of motion for the soft robot is obtained as:
\begin{equation} \label{softrobot_dynamics}
    M(q)\ddot{q} + C(q,\dot{q})\dot{q} + D(q)\dot{q} + K(q) + G(q) = \tau + F_{E}(q,\dot{q})
\end{equation}
where $M \in \mathbb{R}^{n\times n}$, $C \in \mathbb{R}^{n\times n}$, $D(q) \in \mathbb{R}^{n\times n}$ and $K \in \mathbb{R}^{n}$ are the inertia matrix, Coriolis matrix, damping matrix and stiffness matrix respectively. Also, $G(q)\in \mathbb{R}^{n}$ is a vector of gravitational effects (including buoyancy), $\tau \in \mathbb{R}^{n}$ is a vector of actuation torques (further detail can be found in Section \ref{control} in terms of the strategies employed to generate these) and $F_{E}(q,\dot{q})\in \mathbb{R}^{n}$ is a generalized vector of external disturbances including those induced by the interaction with the surrounding fluid.

The present control is postulated in an actuation-agnostic form, leaving the mapping from torque to actuation-specific control input unprescribed. This ensures that the control method is generalizable to a broader spectrum of actuation types (e.g. tendon-driven, fluidic, etc.). A generalized disturbance vector is employed to evaluate the effect of the fluid across the body when either the fluid or body is non-stationary (or both); this takes the form:
\begin{equation} \label{env_model}
    F_{E}(q,\dot{q}) = \sum^{n}_{i=1}\int_{0}^{1} J_{i}^{T}(s,q) \mathcal{F}_{i}(v, \dot{v}) ds 
\end{equation}
where $\mathcal{F}_{i}$ is the linear force acting on the $i$-th element relative to the Cartesian velocity vector $v = [ v_{x}, v_{z} ]^T$. 

\subsection{Hydrodynamic Interactions}\label{hydro_modelling}

Modelling a soft robot in a dense fluid medium introduces additional complexities to the typical dynamics of the soft body. Interaction between the body and the surrounding medium will cause deformations relating to the hydrodynamic parameters of the robot, which are highly nonlinear and fully coupled. Here, a disturbance model is considered inspired by Morison theory \cite{Morison1950}. For simplicity of presentation the dependencies associated with $\mathcal{F}$ are dropped. Firstly, the forces induced by the presence of the fluid are defined as:
\begin{equation}
    \mathcal{F} = \mathcal{F}_{A} + \mathcal{F}_{D}
\end{equation}
where $\mathcal{F}_{A}$ and $\mathcal{F}_{D}$ are the added mass and drag effects respectively. The added mass effect is a product of the body acceleration and an added mass matrix:
\begin{equation}
    \mathcal{F}_{A} = -\mathcal{M}_{A}\dot{v}
\end{equation}
where the coefficients in $\mathcal{M}_{A} \in \mathbb{R}^{2\times2}$ are evaluated empirically in this work, producing \cite{Journee2001}:
\begin{equation} \label{added_mass_matrix}
    \mathcal{M}_{A} = \frac{\pi d_{s}^{2}}{4} L\rho_{f} 
    \begin{bmatrix}
    C_{m,x} & 0 \\
    0 & C_{m,z}
    \end{bmatrix}
\end{equation}
where $d_{s}=0.05$m is the diameter of the segment cross-section, $L=0.3$m is the segment length and $C_{m,x}$, $C_{m,z}$ are inertia coefficients relating to the body added mass where $C_{m} = 1 + C_{a}$. The drag forces are modelled utilising a relative velocity vector (rotated into the local frame of each considered body), considering the normal and tangential velocities at  point, $s$, along the segment, defined as: 
\begin{equation}
    v_{r} = R(s,q) (v - v_{f}) = [ v_{n}, v_{t} ]^T
\end{equation}
where $v_{f} = [ u_{p}, w_{p} ]^T$ is the particle velocity vector in the surge and heave. It follows that:
\begin{equation}
    \mathcal{F}_{D} = - \mathcal{D} v_{r}
\end{equation}
where
\begin{equation} \label{drag_matrix}
    \mathcal{D} = \frac{1}{2}\rho_{f}A_i \begin{bmatrix}
    C_{d}|v_{n}| & 0 \\
    0 & C_{f}|v_{t}|
    \end{bmatrix}
\end{equation}
and $A_i$, $C_{d}$, $C_{f}$ are the incident area to the flow, the drag coefficient of the segment and the frictional coefficient of the segment. It is worth noting that in reality the off-diagonal coefficients of Eq. \ref{added_mass_matrix} \& Eq. \ref{drag_matrix} may be non-zero, however a diagonal structure can be adopted when assuming the body is axisymmetric, implying off-diagonal contributions will be negligible in comparison \cite{FossenBook}. The hydrodynamic properties of the robot will vary according to the morphology of each segment, both due to the nature of the cross-section and the associated segment length; for longer segments with larger cross-sections, an increased area will be subjected to hydrodynamic loading. However, this generalised approach remains consistent and these values can be determined accordingly through experimentation or empirical calculations.


\subsection{Unsteady Wave-Induced Disturbances}

Application of this model relies on knowledge of the fluid particle motions at each point along the soft robot, modelled through Linear Wave Theory (LWT) in this work. As the proposed application is for shallow ocean environments, this simplification was reasonable. According to LWT, a random sea-state is composed of a spectrum of monochromatic components each with a unique wave amplitude, $\mathcal{A}$, period, $T$, and phase offset, $\phi$ \cite{Dean1984}. When superimposed, these components form the sea surface wave elevation, $\eta$, at a specified point in time and space $(x,t)$ according to:
\begin{equation} 
	\label{WaveEquation}
	\zeta(x, t) = \sum_{i=1}^{N} \frac{\mathcal{A}_{i}}{2}cos(k_{i}x - \omega_{i} t + \phi_{i})
	\end{equation}
where $k$, $\omega$ and $\lambda$ represent the wave number, the angular frequency and the wavelength of each component. Knowledge of these characteristics for each wave component facilitates the reconstruction of the local flow field in the whole domain \cite{Dean1984}:
\begin{equation} 
	\label{ParticleVelocityX}
	u_{p}(x, z, t) = \sum_{i=1}^{N} \frac{\pi \mathcal{A}_{i}}{T_{i}}\frac{\cosh k_{i}(z+d)}{\sinh k_{i}d}cos(k_{i}x - \omega_{i} t + \phi_{i})
\end{equation}	
\begin{equation} 
	\label{ParticleVelocityZ}
	w_{p}(x, z, t) = \sum_{i=1}^{N} \frac{\pi \mathcal{A}_{i}}{T_{i}}\frac{\sinh k_{i}(z+d)}{\sinh k_{i}d}sin(k_{i}x - \omega_{i} t + \phi_{i})
\end{equation}
This information can be directly implemented within the theory presented in Section \ref{hydro_modelling} through a function $f_{p}(\mathcal{A},\phi,\omega)$ to simulate the effect of the fluid flow on the body configuration.

\section{Shape Control} \label{control}

\begin{figure}[t!]
    \centering
    \includegraphics[width=\linewidth]{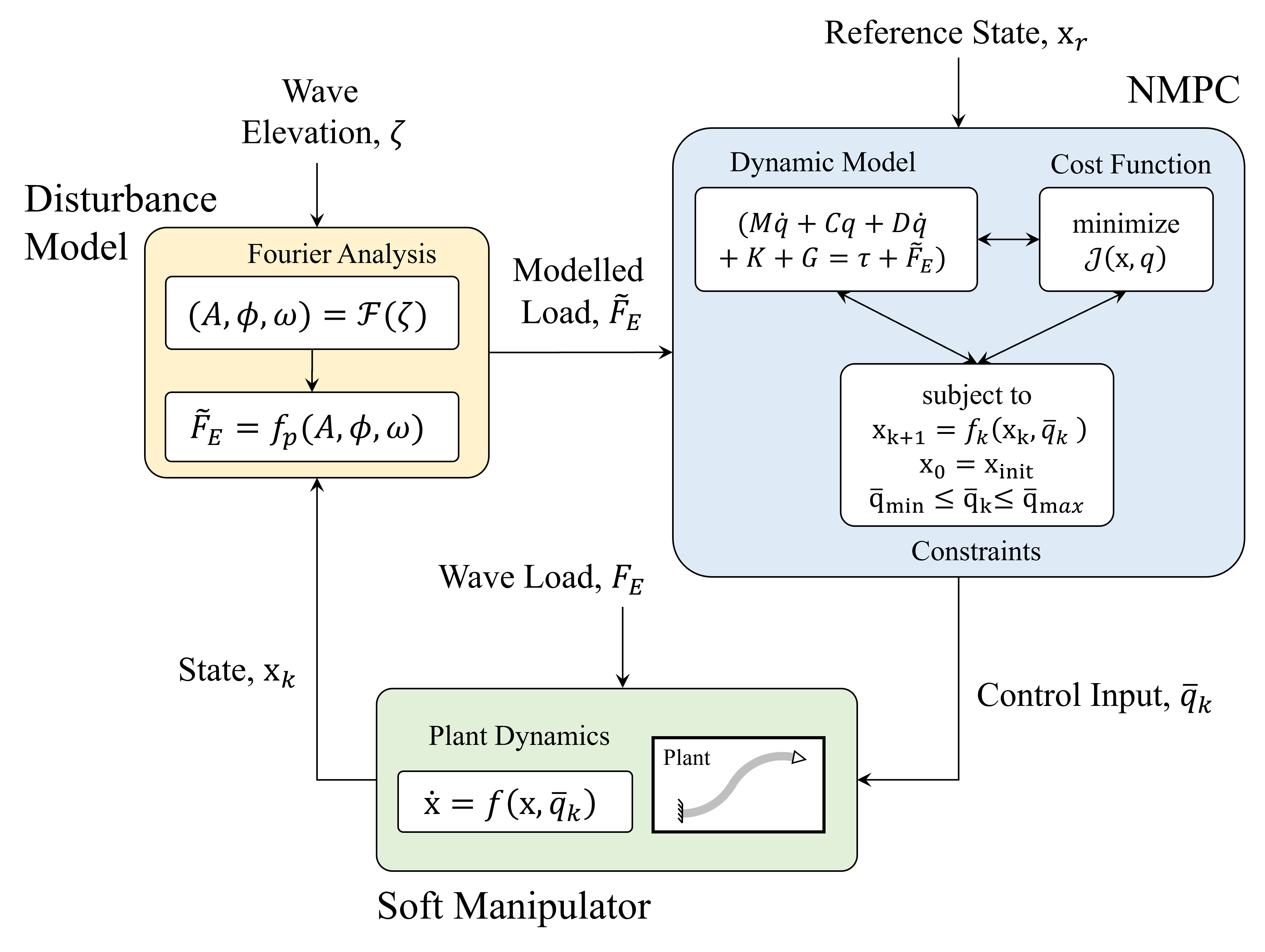}
    \caption{Overview of the control architecture deployed for disturbance rejection; it should be noted dependencies have been dropped here in the model for the sake of space.}
    \label{block_diagram}
    \vspace{-0.2cm}
\end{figure}

\subsection{Model Predictive Control}

Recent work on soft robotic manipulator control has employed MPC to address the high non-linearity of the robotic structure in order to achieve position control. The MPC approach involves calculation of an open-loop control action by considering the system state, a goal state and a specified cost function before applying it over a receding horizon. Subsequently, the control sequence is re-calculated and optimised relative to the cost function, thus passing from open to closed loop \cite{WangBook}. The asset of MPC lies in the ability to handle constraints as well as to account for the system future dynamic response over a short time-horizon. In the case presented here, this latter quality is exploited to incorporate impending hydrodynamic disturbances within the future sequence of control actions (Fig. \ref{block_diagram}), aiming to adapt the optimal control output to minimise environmental disturbances. 

The assumption is made that the initial end-effector state of the robot $\mathbf{x}_{init} \in \mathbb{R}^2$ is known and a reference state is defined at each timestep $k$ such that $\mathbf{x}_{r,k} = [ x_{r,k}, z_{r,k} ]^T \in \mathbb{R}^{2}$. Formulating the problem as an optimisation over the state and control trajectory, the goal of the MPC is to evaluate a set of control inputs $\Bar{q}$ which minimise a specified cost function. These control inputs are then passed to the lower level kinematic controller given in Eq. \ref{FF_PD} to generate an applied actuation torque vector. To this end we solve the following optimal control problem:
\begin{equation}
\begin{gathered}
\label{cost_func}
    \underset{({\Bar{q}}_{0},\dots,{\Bar{q}}_{K})}{\arg\min} \mathcal{J} = \sum_{k=0}^{K-1} (\mathbf{x}_{k}-\mathbf{x}_{r})^{T}Q(\mathbf{x}_{k}-\mathbf{x}_{r}) \\ + \Delta \Bar{q}_{k}^T R\Delta \Bar{q}_{k}
    \end{gathered}
\end{equation}
\begin{equation*}
    \begin{aligned}
    s.t. \quad & \mathbf{x}_{k+1} = f(\mathbf{x}_{k},\Bar{q}_{k}) \\
    & \mathbf{x}_{0} = \mathbf{x}_{init} \\
    & \Bar{q}_{min} \leq \Bar{q}_{k} \leq \Bar{q}_{max} \\
    \end{aligned}
\end{equation*}

where the function $f(\mathbf{x}_{k},\Bar{q}_{k})$ describes the state evolution defined by the system dynamics. Also, $\mathbf{x}_{k}$ is the current state at each instance along the control horizon and $Q\in\mathbb{R}^{n\times n}$, $R\in\mathbb{R}^{n\times n}$ are positive definite weight matrices on the state and control respectively. The control input takes the form of a commanded joint angle configuration $\Bar{q}$ and $\Delta\Bar{q}$ represents the intermediate step; these terms are included within Eq. \ref{cost_func} to prevent large step changes and minimise the required effort. 

As we consider a generalised actuation, we only place constraints on the desired joint angle rather than the torque directly. The MPC constraints can be easily adapted for this to include consideration for effects such as saturation. The optimised joint-angle sequence is the direct input to Eq. \ref{FF_PD} to actively adapt the robot configuration to the disturbance. Although the control inputs are applied in the joint-space, the optimisation is performed with reference to the task-space in order to operate around a position rather than a configuration. Solving for the subsequent state $\mathbf{x}_{k+1}$ was performed through numerical integration over the discrete time interval $\Delta t = 0.1$s using a fifth-order variable step Runge-Kutta method. For our analysis, the prediction horizon was defined as $t_{p} = 15\Delta t$. 

\begin{figure}[t!]
    \centering
    \includegraphics[width=0.95\linewidth]{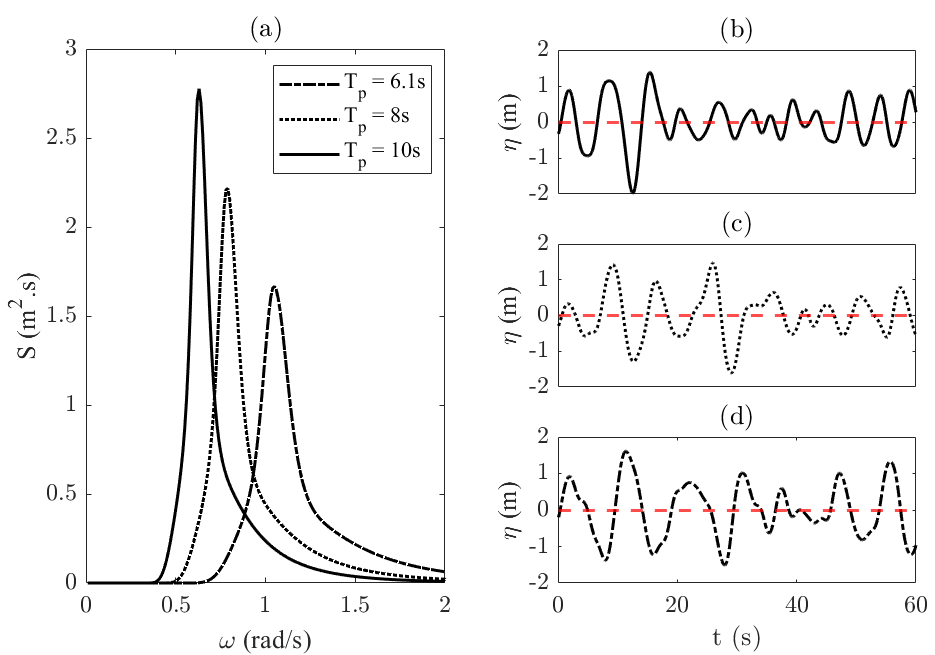}
    \caption{The various wave spectra selected for testing, all based on real-world data collected using wave buoys, showing (a) the JONSWAP curve for each spectra and the temporal representation wave cases (b) W1, (c) W2 and (d) W3 (each scaled to $H_s = 3m$).}
    \label{spectra}
\end{figure}
\begin{table}[t!]
\caption{Spectral characteristics of the three different wave spectra tested with the control framework.}
\label{wave_cases}
\centering
\renewcommand{\arraystretch}{1.3}
\begin{tabular}{|c|c|c|}
\cline{1-3}
Wave Reference & Peak Period (s) & Significant Height (m) \\ \hline \hline 
W1             & 6.1                        & 0.5-3 (0.5m intervals) \\ \hline 
W2             & 8                       & 0.5-3 (0.5m intervals)  \\ \hline 
W3             & 10                       & 0.5-3 (0.5m intervals)     \\ \hline 
\end{tabular}
\end{table}

\subsection{Model-Based Kinematic Control}\label{baseline_controller}

To present an informative comparison, a kinematic controller was utilised as a baseline controller. For convenience we elect to formulate the manipulator control in the task-space, which for the soft robot under consideration relates to the position of the end-effector. The kinematic approximation is well covered in literature \cite{DellaSantina2021} and we therefore just present the control strategy adopted for the baseline controller in this work. It follows that the actuation torque, $\tau$, is generated by application of a feed-forward with PD control law such that
\begin{equation} \label{FF_PD}
    \tau(\Bar{q},q,\dot{q}) = K(\Bar{q}) + G(\Bar{q}) + \alpha(\Bar{q}-q) - \beta(\dot{q})
\end{equation}
where $\alpha$ and $\beta$ are two gain matrices on the proportional and derivative actions, respectively. The commanded joint-angle configuration, $\Bar{q}$, is determined by applying a kinematic control loop as a planning strategy where
\begin{equation}
    \dot{\Bar{q}} = J^{+}(\Bar{q})(K_{e}(\Bar{x}-h(\Bar{q})))
\end{equation}
where $K_{e}$, $\Bar{x}$ and $h(\Bar{q})$ are a gain, the desired task-space set-point and the forward kinematics of the joint configuration, respectively, and $J^{+}(q)$ is the Moore-Penrose pseudo-inverse of $J$.

\section{Simulation Results}

\begin{table}[t!]
\caption{Desired end-effector positions the controller is tasked with regulating in Section \ref{posture_regulation_results}, given in Cartesian co-ordinates.}
\label{poses}
\centering
\renewcommand{\arraystretch}{1.3}
\begin{tabular}{|c|c|}
\cline{1-2}
Pose Reference & End-Effector Cartesian Position (x,z) \\ \hline \hline 
P1  & (0.3,-3.7)m \\ \hline 
P2  & (0.5,-3.7)m \\ \hline 
P3  & (0.7,-3.7)m \\ \hline 
P4  & (0.3,-4.3)m \\ \hline 
P5  & (0.5,-4.3)m \\ \hline 
P6  & (0.7,-4.3)m \\ \hline 
\end{tabular}
\end{table}

The manipulator is modelled with three individually actuated segments which are all assumed to be identical with regards to their physical properties and hydrodynamic characteristics. The simulated scenario was configured to replicate the conditions a vehicle-manipulator system may encounter. Real-world data was sourced to emulate a typical shallow-water environment, collected using a wave buoy located within the Moray Firth, an inlet off the cost of Inverness, Scotland ($57^{\circ}57'.99$N, $003^{\circ}19'.99$W) through the online repository of the Centre For Environment Fisheries and Aquaculture Science (Cefas) \cite{cefas}. Three different wave spectra were selected as depicted in Fig. \ref{spectra}, purposefully choosing varying spectral peak periods and scaling wave height (listed in Table \ref{wave_cases}) to analyse any direct effect on the controller performance. For each test performed, a 60s temporal segment was analysed with the manipulator situated at an operating depth of $z=4$m, implying wave disturbances will significantly influence the robot dynamic behaviour. It should be noted that the estimated wave disturbances considered by the control are those defined by Eq. \ref{env_model} with added white gaussian noise (SNR = 20); the assumption here is that reasonably accurate spectral information can be obtained using a method similar to \cite{WalkerIROS}.

\begin{figure}[t!]
    \centering
    \includegraphics[width=0.97\linewidth]{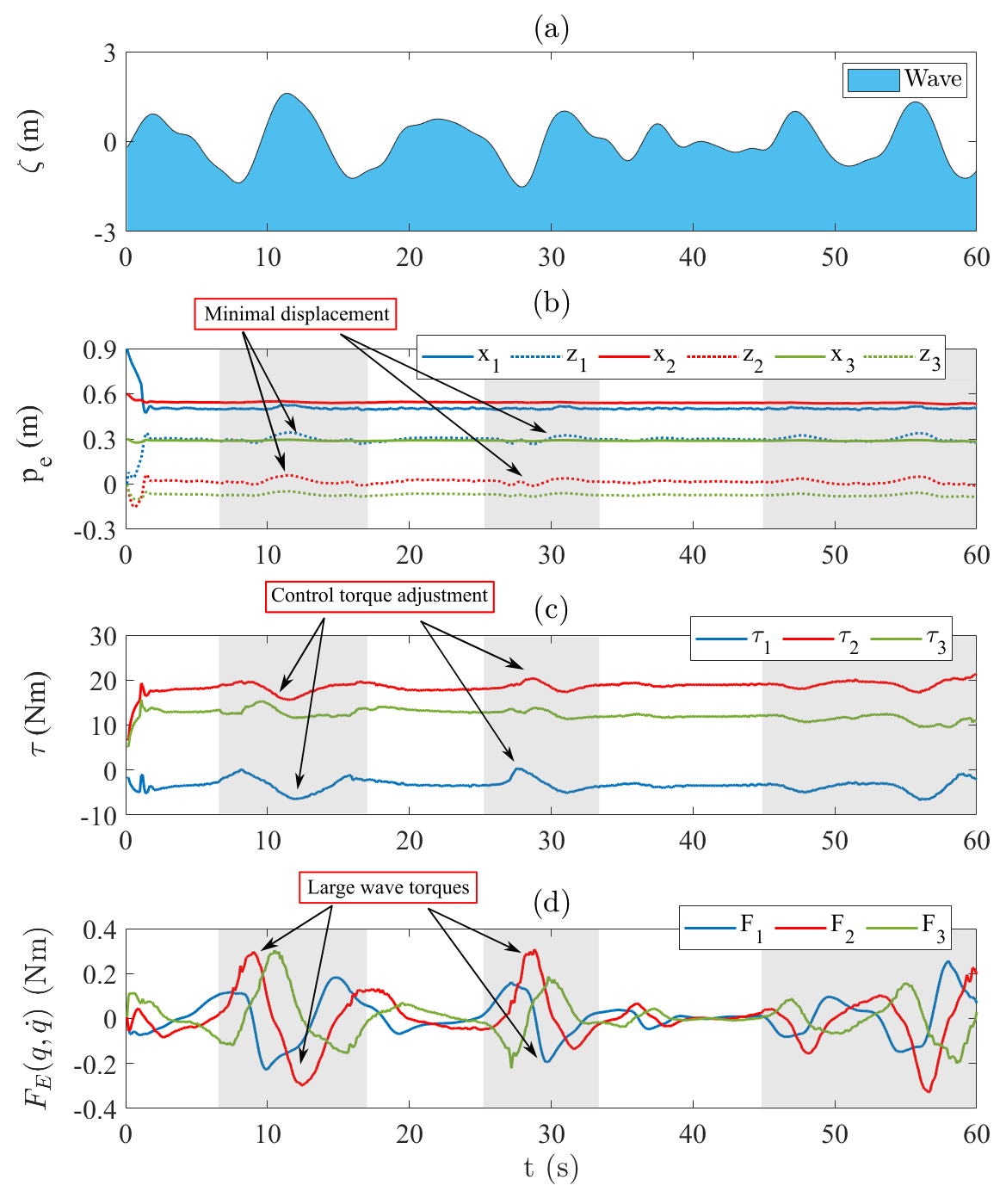}
    \caption{When subjected to the (a) wave train, the evolution of the (b) segment tip positions with respect to the base at (0m, 0m), (c) actuation torques and (d) wave loading torques are analysed. Shown are the evolution's for case W3 and highlighted in grey are key points of note where large disturbances are impacting the manipulator.}
    \label{example_case}
    \vspace{-0.2cm}
\end{figure}

\subsection{Disturbance Rejection Performance}
\label{posture_regulation_results}
We assess the overall controller capability to compensate for wave disturbances by undertaking set-point regulation tasks over the cases previously mentioned. To illustrate the behaviour of the control when considering disturbances across the temporal interval, Fig. \ref{example_case} shows the evolution of the position, control torques and wave loading for each of the three segments under the wave train depicted in Fig. \ref{example_case}(a). Subscripts 1, 2 and 3 refer to the base, middle and end-effector segments respectively. With reference to the hydrodynamic torques, Fig. \ref{example_case}(d) shows the independent fluid loading across each segment. As can be seen, the actuation torque applied by the controller varies at each wave loading point of inflection, showing an active effort to compensate for the disturbance and maintain the end-effector position steady. 

\begin{figure}[!t]
    \centering
    \includegraphics[width=0.9\linewidth]{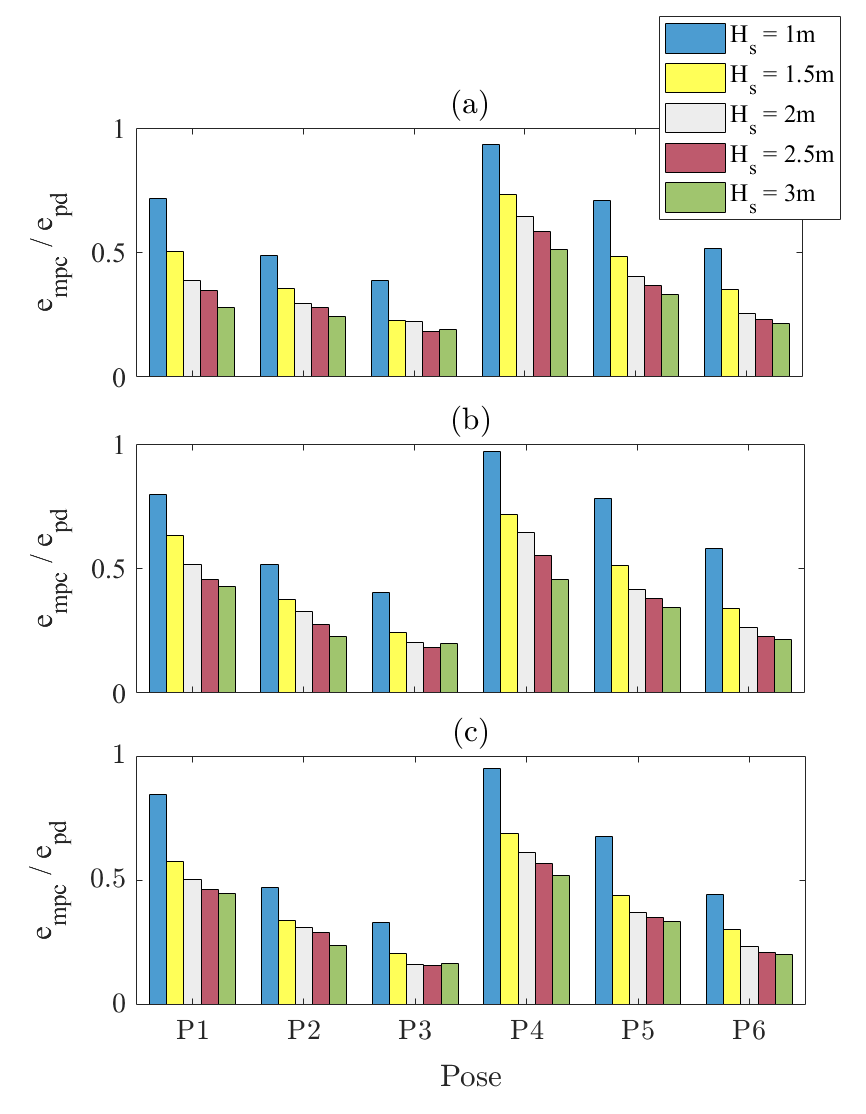}
    \caption{Variation in error between the MPC strategy and the baseline feedforward + PD strategy, represented as an RMSE ratio. Shown are the results for cases (a) W1 (b) W2 and (c) W3.}
    \label{error_bar}
    \vspace{-0.2cm}
\end{figure}

\begin{figure*}[t!]
\captionsetup[subfigure]{justification=centering}
    \centering
    \begin{subfigure}[t]{0.32\textwidth}
        \centering
        \includegraphics[width=0.9\textwidth]{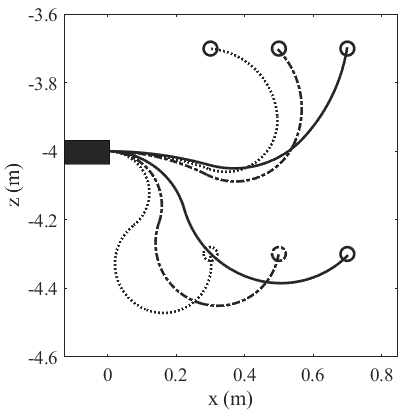}
        \caption{All three segments controllable.}
    \end{subfigure}%
    ~ 
    \begin{subfigure}[t]{0.315\textwidth}
        \centering
        \includegraphics[width=0.9\textwidth]{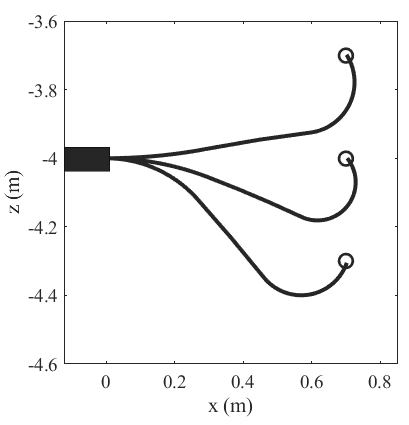}
        \caption{Second (mid) segment uncontrollable.}
    \end{subfigure}
    ~ 
    \begin{subfigure}[t]{0.315\textwidth}
        \centering
        \includegraphics[width=0.9\textwidth]{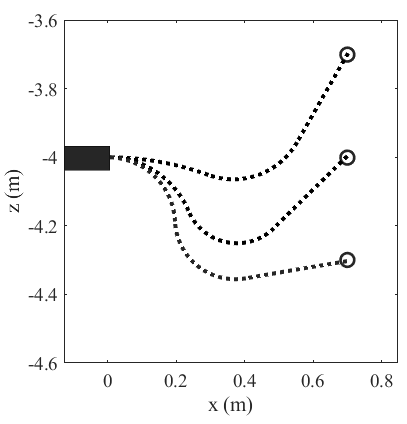}
        \caption{Third (tip) segment uncontrollable.}
    \end{subfigure}
    \caption{Set-point regulation validation of the control with varying actuation capability, showing poses at different end-effector depths and degrees of segment curvature.}
    \label{various_setpoints}
    \vspace{-0.2cm}
\end{figure*}


To quantify the performance against the baseline controller specified in Section \ref{baseline_controller}, identical tasks were undertaken and the RMSE of both strategies compared. These extensive results are displayed in Fig. \ref{error_bar} by means of a non-dimensional error ratio, where $e_{mpc}$ and $e_{pd}$ refer to the MPC and Kinematic control respectively and $e_{mpc}/e_{pd} = 1$ relates to equal performance. These results highlight the superior performances of the MPC against the baseline controller across all cases, showing substantial improvement in end-effector control under disturbances. An intuitive observation is that the reduction in error becomes more significant as the magnitude of disturbances grow, an expected outcome as the baseline control performance will deteriorate at a faster rate with growing disturbances. This largely owes to the ability of the MPC to consider modelled disturbances explicitly within the control architecture. Likewise, the peak period of the wave appears to have a significant effect on the MPC performance, particularly when the manipulator is attempting to maintain an end-effector position close to the base, i.e. when all segments undergo high curvatures. This aligns with the expectation that elongation of the fluid trajectories beneath the surface \cite{Dean1984} has a direct influence on regulation performance.

The largest reductions in error for each period across all poses and wave heights were $81.67\%$, $81.60\%$ and $83.78\%$ for cases W1, W2 and W3 respectively, whilst the lowest reductions were $6.48\%$, $2.82\%$ and $5.12\%$. Interestingly, these all referred to the same pose, the larger reductions relating to pose P3 and the lowest to P4. The inference here is when the soft robot experiences higher curvatures, the improvement in set-point regulation is less drastic due to the fact that the body is experiencing a lower variation in flow, therefore a lower disturbance torque. A similar cause is that higher depth will also imply lower disturbance torques, so the variation in performance is expected to be lower. Ocean wave effects decay exponentially with depth \cite{Dean1984}, so it is unsurprising that a clear difference can be seen between all set-points at $z=-3.7$m compared to $z=-4.3$m.


\subsection{Actuation Failure Mitigation}

An interesting aspect of deploying MPC for regulation tasks is the ability to adapt and remain functional under variations to system dynamics. As a plant model is exploited in the optimisation of the control inputs, adaptations to adverse scenarios become simpler and more accessible. The example we consider in this instance is one of the actuated segments "failing", i.e. can no longer be actuated and becomes passive, meaning the controller must compensate for by adapting the actuation of the other segments. 

Considering this, simulations were performed emulating instances where a segment could not be actuated directly and the end-effector position was purely controlled by the configuration of the other segments. As displayed in Fig. \ref{various_setpoints}(b)(c), the controller was able to reconfigure the soft robot to successfully reach the desired end-effector position, both when the second and third segment were unactuated and could not be controlled (independent of each other). This is highly advantageous when considering the proposed application and operational environment of these systems. The potential of the manipulator sustaining damage is increased during these operations, so if an instance arose where the actuation failed, using advanced control like the MPC proposed here mitigates the risk of mission failure. The limitation in this instance is that the workspace is reduced, however the ability to still partially operate is highly desirable when operating in extreme environments. Generally, it is beneficial to maintain some functionality and attempt to continue operating rather than abandoning the mission completely, particularly from a costs incurred perspective.

\begin{figure}[t!]
    \centering
    \includegraphics[width=0.9\linewidth]{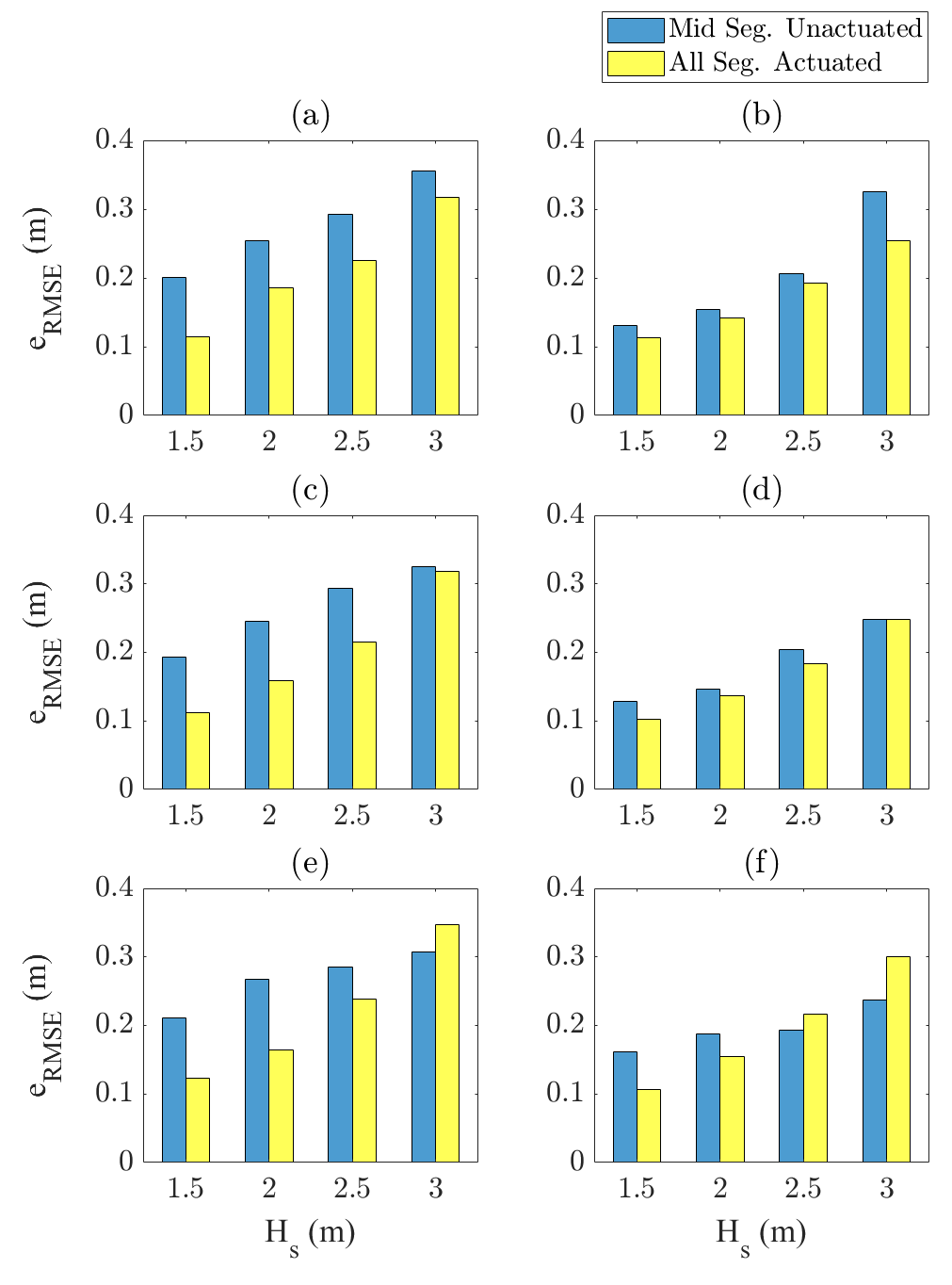}
    \caption{RMSE of the soft robot end-effector in the presence of a mid-section actuation failure, in comparison to a fully actuated soft robot. Shown are the results for cases (a)(b) W1, (c)(d) W2 and (e)(f) W3 when attempting to regulate the end-effector pose (a)(c)(e) P3 and (b)(d)(f) P6. }
    \label{actuation_failure_rmse}
    \vspace{-0.2cm}
\end{figure}

With regards to quantitative performance, the results presented in Fig. \ref{actuation_failure_rmse} provide further evidence of the controller ability to still regulate the end-effector position with minimal increase in RMSE. Two poses were tested, P3 and P6, as these cases contrast in operational depths with the robot fully extended, so the change in performance was of interest. Likewise, significant wave heights in the range of $H_{s} = 1.5-3$m were selected to restrict the analysis to instances of large disturbances. Across all spectra, only a $25.61\%$ increase in RMSE was witnessed, which given that the robot has lost 1/3 of it's manoeuvring ability is quite remarkable. Still comparing the partially actuated case to the fully actuated case, the wave with the largest peak period (case S3) showed the lowest disparity, recording a $23.17\%$ increase. Although the largest wave is expected to induce the highest torques, this result could point to the higher frequency waves causing a passive excitation of the middle segment, thus driving the error higher than when the segment is controlled. It should be noted that this difference was marginal and differed $<4\%$ compared to the other spectra. Interestingly, there were some instances where the fully actuated robot actually recorded larger RMSE than the partially actuated robot. However, a key observation is that these cases exclusively relate to waves with the largest significant wave height tested. The inference here is that when the wave becomes substantially larger than the robot body length ($\approx$ 3x larger), the control has difficulty regulating the end-effector position regardless of actuation capability. It should be noted however that the values recorded are still a significant improvement in comparison to standard control techniques which do not consider a time-history of the wave disturbances within the control, as demonstrated in Fig. \ref{error_bar}.

\subsection{Trajectory Tracking}

\begin{figure}[t!]
    \centering
    \includegraphics[width=0.9\linewidth]{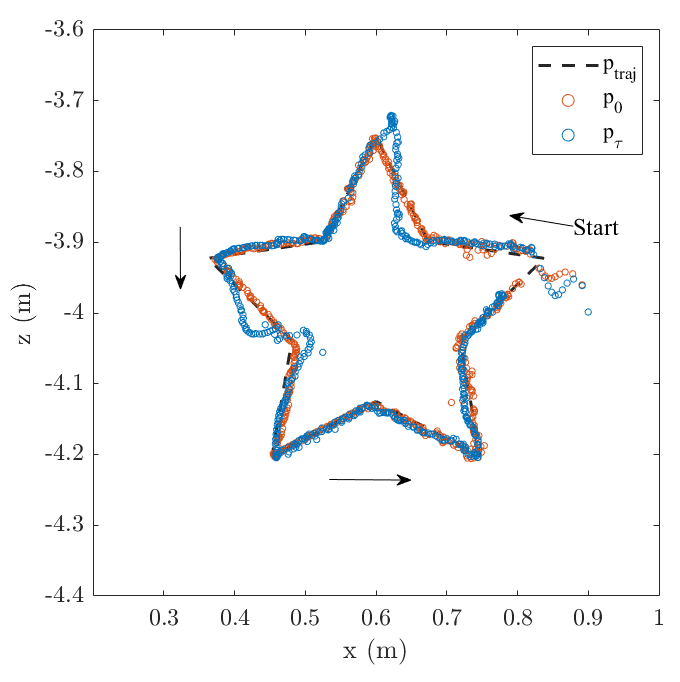}
    \caption{Trajectory tracking task for a star path, initial condition fully extended and subject to the wave elevation in Fig. \ref{example_case} (case W3, $H_{s}=3$m), showing the end-effector position with respect to the base at (0m, 0m).  Subscripts $\tau$ and $0$ refer to cases under wave disturbances and neglecting disturbances respectively. }
    \label{star_trajectory}
    \vspace{-0.2cm}
\end{figure}

All tests in the previous section involve defining a constant target position which the controller attempts to regulate under disturbances. An intuitive further extension is to instead define a trajectory which the end-effector must follow with minimal fluctuations while subject to wave loading. With respect to this, a star trajectory in the task space was prescribed for the controller to attempt to track with the end-effector. All three wave spectra were considered for this demonstration (Fig. \ref{spectra}) with significant wave height $H_{s} = 3$m. A visual representation of the task for case W3 is given in Fig. \ref{star_trajectory}(d), which brings evidence of minimal discrepancy regardless of the magnitude of the wave disturbance. Evaluation of controller performance across the three spectra over a $60$s time interval yields RMSE of $0.2491$m, $0.2944$m and $0.2821$m for cases W1, W2, and W3 respectively. This shows relatively consistent performance irrespective of wave period when tracking a trajectory, a markedly distinct behaviour from other underwater floating-base systems \cite{Walker2021_RAL} such as ROVs. This is justifiable by considering that, at such wave lengths, drag-based forcing will be predominant and acting on the manipulator over timescales which are much larger than the control timescale.


\section{Conclusions}

A predictive control strategy was successfully deployed in simulation on a multi-segment soft robot for rejection of active time-varying disturbances. The controller is tested under realistic scenarios which underwater manipulators commonly encounter, i.e. set-point regulation, trajectory tracking and fault compensation. The modelled disturbances were incorporated within the control through a NMPC strategy, displaying significant performance improvements in comparison to a model-based kinematic control strategy, with up to $\approx84\%$ reduction in error witnessed. Successful trajectory tracking tests under wave disturbances were also observed, showing constrained positional error throughout the tracking task irrespective of wave conditions. Finally, robustness to actuation failure was demonstrated, whereby the controller was still able to attain a desired end-effector position when only controlling 2 of the 3 segments. This is promising with regards to operating in extreme environments, as the implication is certain tasks could still be completed under partial system failure. 

With reference to the points raised in the introduction regarding difficulties of operating in a marine environment, MPC shows potential for improving the control of underwater manipulators by accounting for disturbances explicitly within the control sequence optimisation, while accounting for system dynamics to minimize such disturbances. The overall performances of the MPC are found to scale positively with wave heights, Fig. \ref{error_bar}, further supporting the use of MPC as a control solution for real-world operation of soft manipulators.

\addtolength{\textheight}{-12cm}   






\begin{thebibliography}{99}






\bibitem{Aracri2021} S. Aracri, F. Giorgio-Serchi, G. Suaria, M. E. Sayed, M. P. Nemitz, S. Mahon and Adam A. Stokes, "Soft Robots for Ocean Exploration and Offshore Operations: A Perspective", \emph{Soft Robotics}, vol. 8, no. 6, pp. 625-639, 06 Dec. 2021.

\bibitem{Trivedi2008} D. Trivedi, C. D. Rahn, W. M. Kier and I. D. Walker, "Soft Robotics: Biological Inspiration, State of the Art, and Future Research", \emph{Applied Bionics and Biomechanics}, vol. 5, pp. 99-117, 2008.

\bibitem{Renda2014} F. Renda, M. Giorelli, M. Calisti, M. Cianchetti and C. Laschi, "Dynamic Model of a Multibending Soft Robot Arm Driven by Cables", \emph{IEEE Transactions on Robotics}, vol. 30, no. 5, Oct. 2014.

\bibitem{Galloway2016} K. C. Galloway, K. P. Becker, B. Phillips, J. Kirby, S. Licht, D. Tchernov, R. J. Wood, and D. F. Gruber, "Soft Robotic Grippers for Biological Sampling on Deep Reefs", \emph{Soft Robotics} vol. 3, no. 1, March 2016.

\bibitem{Gong2021} Z. Gong, X. Fang, X. Chen, J. Cheng, Z Xie, J. Liu, B. Chen, H. Yang, S. Kong, Y. Hao, T. Wang, J. Yu and L. Wen, "A soft manipulator for efficient delicate grasping in shallow water: Modeling, control, and real-world experiments", \emph{
The Intl. Jnl. of Rob. Res.}, vol. 4, no. 1, pp. 499-469, 2021.

\bibitem{Calisti2017} M. Calisti, G. Picardi and C. Laschi, "Fundamentals of soft robot locomotion", \emph{Jnl. of the Royal Soc. Interface}, vol. 14, no. 130, 2017. 


\bibitem{GiorgioSerchi2017} F. Giorgio-Serchi and G.D. Weymouth, "Underwater Soft Robotics, the Benefit of Body-Shape Variations in Aquatic Propulsion", Soft Robotics: Trends, Applications and Challenges, Springer International Publishing, 37--46, 2017.


\bibitem{Armanini2021} C. Armanini, M. Farman, M. Calisti, F. Giorgio-Serchi, C. Stefanini and F. Renda, "Flagellate Underwater Robotics at Macroscale: Design, Modeling and Characterization", \emph{IEEE Transactions on Robotics}, vol. 38, no. 2, pp. 731-747, July 2021.

\bibitem{Dian2022} S. Dian, Y. Zhu, G. Xiang, C. Ma, J. Liu and R. Guo, "A Novel Disturbance-Rejection Control Framework for Cable-Driven Continuum Robots With Improved State Parameterizations", \emph{IEEE Access}, vol. 10, pp. 91545-91556, 2022.

\bibitem{DellaSantina2021} C. Della Santina, C. Duriez and D. Rus, "Model Based Control of Soft Robots: A Survey of the State of the Art and Open Challenges", \emph{preprint arXiv:2110.01358}, 2021.

\bibitem{Best2016} C. M. Best, M. T. Gillespie, P. Hyatt, L. Rupert, V. Sherrod and M. D. Killpack, "A New Soft Robot Control Method: Using Model Predictive Control for a Pneumatically Actuated Humanoid", \emph{IEEE Robotics \& Automation Magazine}, vol. 23, no. 3, pp. 75 - 84, Sept. 2016.

\bibitem{Bruder2019} D. Bruder, B. Gillespie, C. D. Remy and R. Vasudevan, "Modeling and Control of Soft Robots Using the Koopman Operator and Model Predictive Control", \emph{in Proc. of Robotics: Science and Systems}, Freiburg im Breisgau, Germany, 22-26 June, 2019.

\bibitem{Hyatt2020} P. Hyatt, C. C. Johnson and M. D. Killpack, "Model Reference Predictive Adaptive
Control for Large-Scale Soft Robots", \emph{Frontiers in Robotics and AI 7}, 05 Oct. 2020.

\bibitem{Fernandez2017} D. C. Fernandez and G. A. Hollinger, “Model Predictive Control for Underwater Robots in Ocean Waves”, \emph{IEEE Robotics and Automation Letters}, vol. 2, no. 1, pp. 88–95, 2017.

\bibitem{Walker2021_RAL} K. L. Walker, R. Gabl, S. Aracri, Y. Cao, A. A Stokes, A. Kiprakis and F. Giorgio-Serchi, "Experimental validation of wave induced disturbances for predictive station keeping of a remotely operated vehicle", \emph{IEEE Robotics and Automation Letters}, vol. 6, no. 3, pp. 5421-5428, 2021.

\bibitem{Loo2019} J. Loo, C. Tan and S. Nurzaman, "H-infinity based extended kalman filter for state estimation in highly non-linear soft robotic system", \emph{Proc. American Control Conference}, Philadelphia, PA, USA, pp. 5154-5160, 10-12 July 2019.

\bibitem{Hughes2021} J. Hughes, F. Stella, C. Della Santina and D. Rus, "Sensing Soft Robot Shape Using IMUs: An Experimental Investigation", \emph{Proc. International Symposium on Experimental Robotics}, pp. 543-552, 2021.

\bibitem{Hu2023} D. Hu, F. Giorgio-Serchi, S. Zhang and Y. Yang, "Stretchable e-skin and transformer enable high-resolution morphological reconstruction for soft robots", \emph{Nature Machine Intelligence}, pp. 2522-5839, 2023.

\bibitem{Webster2010} R. J. Webster III and B. A. Jones, "Design and kinematic modeling of constant curvature continuum robots: A review", \emph{The International Journal of Robotics Research}, vol. 29, no. 13, pp. 1661-1683, 2010.

\bibitem{Morison1950} J. R. Morison, M. P. O'Brien, J. W. Johnson and S. A. Schaaf, "The force exerted by surface waves on piles",  J. Pet. Technol., vol. 2, no. 5, pp 149-154, May 1950.

\bibitem{Journee2001} J. M. J. Journ\'{e}e and W. W. Massie, "Offshore Hydromechanics", \emph{First Edition}, Delft University of Technology, Jan. 2001.

\bibitem{FossenBook} T. I. Fossen, “Guidance and control of ocean vehicles,” \emph{Wiley}, 1994.

\bibitem{Dean1984} R. G. Dean and R. A. Dalrymple, “Water wave mechanics for engineers and scientists.”, 2nd Edition, \emph{World Scientific}, 1984.

\bibitem{WangBook} W. Liuping, "Model Predictive Control System Design and Implementation Using MATLAB", \emph{Springer}, 2009.

\bibitem{cefas} \emph{Wave Spectra Dataset: Feb-May 2021, Moray Firth WaveNet Site}, Centre For Environment Fisheries and Aquaculture Science (Cefas) Online Repository, downloaded July 2022. [Online]. Available: https://www.cefas.co.uk/data-and-publications/wavenet/.

\bibitem{WalkerIROS} K. L. Walker and F. Giorgio-Serchi, "Disturbance Preview for Nonlinear Model Predictive Trajectory Tracking of Underwater Vehicles in Wave Dominated Environments", \emph{Proc. IEEE International Conference on Intelligent Robots and Systems (IROS)}, Detroit, MI, USA, 01-05 Oct. 2023.

\end{thebibliography}
\end{document}